\documentclass[preprint,review,3p,10pt,times]{elsarticle}

\usepackage{natbib}
\usepackage{multirow} 
\usepackage[labelfont=bf]{caption}
\usepackage{booktabs} 
\usepackage{xcolor}
\usepackage{algorithm}
\usepackage{algorithmic}

\usepackage{amssymb}
\usepackage{amsmath}

\usepackage{hyperref}   
\usepackage{booktabs}
\usepackage{multirow}
\usepackage{xcolor}
\usepackage{graphicx} 

\usepackage{makecell}
\usepackage{graphicx} 
\usepackage{booktabs}
\usepackage{multirow}

\usepackage{amssymb}    
\usepackage{tabularx}   

\journal{Pattern Recognition}

\begin{document}

\begin{frontmatter}


  \title{FSCM: Frequency-Enhanced Spatial–Spectral Coupled Mamba for Infrared Hyperspectral Image Colorization }

\author[label1]{Tingting Liu}

\author[label1]{Yuan Liu}

\author[label2]{Guiping Chen}

\author[label1]{Xiubao Sui\corref{cor1}}

\author[label3]{Qian Chen}

  \affiliation[label1]{organization={School of Electronic and Optical Engineering, Nanjing University of Science and Technology},
    city={Nanjing},
    postcode={210094},
    country={China}}

  \affiliation[label2]{organization={School of Mechanical Engineering, University of Science and Technology Beijing},
    city={Beijing},
    postcode={100083},
    country={China}}

  \affiliation[label3]{organization={School of Instrument and Electronics, North University of China},
    city={Taiyuan},
    postcode={030051},
    country={China}}

  \cortext[cor1]{Corresponding authors: Xiubao Sui }

  \begin{abstract}
Thermal infrared imaging is robust to illumination variations and smoke interference, making it important for all-weather perception. However, the lack of natural color and fine texture limits target recognition, human visual interpretation, and the transfer of visible-light models. Existing infrared colorization methods mainly rely on single-band images, where insufficient spectral cues may lead to structural distortion and semantic confusion. Although infrared hyperspectral images provide rich spectral responses and material information, existing single-band frameworks remain limited in modeling spatial-spectral coupling and weak texture details. To address these issues, this paper presents FSCM, a spectral-information-guided GAN framework. Within FSCM, a frequency-enhanced spatial-spectral state-space generator composed of cascaded FSB units is constructed. Each FSB integrates three complementary components: state-space modeling captures global spatial–spectral dependencies; the frequency enhancement module (FEM) combines multi-level wavelet decomposition and Fourier gating to recover structural contours, directional high-frequency details, and global frequency responses; and the dual-stream hybrid gating module (DGM) integrates deformation-aware sampling with sparse attention to enhance effective local structures and suppress background interference. Additionally, an online semantic segmentation-guided loss is introduced to constrain the generated results, improving semantic consistency in complex road scenes. Experiments show that FSCM outperforms existing infrared colorization methods in visual quality and semantic fidelity.

  \end{abstract}


\end{frontmatter}

\section{Introduction}
\label{sec1}
Thermal infrared (TIR) imaging captures radiation emitted by objects, enabling applications like nighttime surveillance and security inspection \cite{TIR}. However, infrared images are typically grayscale and lack color and texture information. This not only weakens human perception of object category, material, and scene semantics, but also limits the transfer of visual models trained on visible-light images \cite{FoalGAN}. Therefore, converting infrared images into colorized visible images with a natural appearance is of practical importance for improving human–machine interaction and downstream perception tasks \cite{Morngan}.

Infrared image colorization is a severely ill-posed cross-domain learning problem. Infrared grayscale imagery mainly reflects thermal radiation intensity, while reflectance- and illumination-related cues that determine visible colors are unavailable \cite{LKATGAN}. Consequently, the same infrared observation may admit multiple plausible RGB appearances when mapped to the visible domain~\cite{CCLGAN}. Compared with conventional colorization of visible-range grayscale images, infrared-to-visible translation is affected by systematic differences in imaging physics, contrast distribution, and detail formation, so purely pixel-wise regression or style-transfer strategies often fail to jointly preserve geometric structure, semantic consistency, and color naturalness. Moreover, thermal diffusion and point-spread effects can blur edges and suppress local contrast, causing semantic adhesion between small targets and the background and further amplifying ambiguity during colorization.

\begin{figure}[t]
  \centering
  \includegraphics[width=0.56\linewidth]{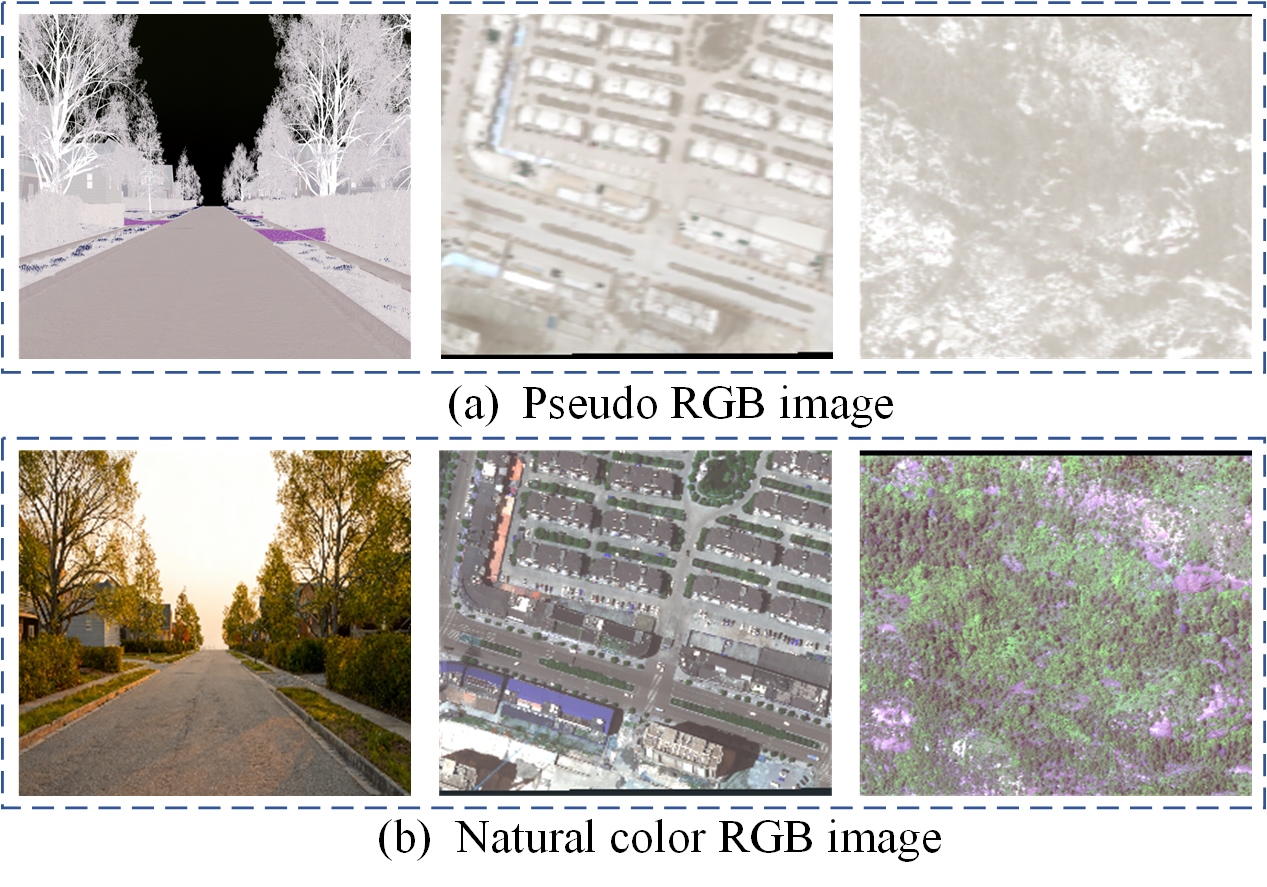}
  \caption{Comparison between an infrared hyperspectral pseudo-RGB image and a natural color image. The pseudo-RGB image is generated by mapping the 1/3, 2/3, and final spectral bands to the B, G, and R channels, respectively. (a) The pseudo-RGB image. (b) The corresponding natural color image.}
  \label{fig:frist_Pseudo_RGB}

  \vspace{-0.45cm}
\end{figure}

In contrast, infrared hyperspectral imagery samples the same scene over multiple narrow bands, encoding temperature distribution together with material-dependent emissive spectral signatures, thereby providing richer cues for recognition, interpretation, and color recovery. Nevertheless, hyperspectral data are high-dimensional and strongly correlated across bands, which are not well handled by methods designed for single-band inputs. Under limited training data, hyperspectral representations are also susceptible to the Hughes phenomenon, leading to degraded performance \cite{BGSR}. In practice, three-band compositing is often used for pseudo-RGB visualization, but it typically produces monotonous tones and weak target color separability (as illustrated in \autoref{fig:frist_Pseudo_RGB}).

Early infrared colorization mainly relied on hand-crafted schemes (e.g., look-up tables, histogram matching, and multi-source fusion), which depend on expert priors and generalize poorly \cite{Video_colorization_TCSVT, Video_colorization_TCSVT_1,Video_colorization_TCSVT_2,Thermal_Infrared_TCSVT}. With the development of deep learning, CNN- and GAN-based methods have become dominant: CNNs learn infrared-to-visible mapping via supervised regression, while GANs add adversarial constraints to enhance realism. Despite progress, three limitations persist. First, CNNs are limited by local receptive fields, whereas Transformers capture global interactions via self-attention at the cost of quadratic token complexity \cite{MBNet}, making accuracy–efficiency trade-offs difficult. Second, feature learning is often focused on the spatial domain, and frequency-domain priors and high-frequency textures are underused, leading to blur and artifacts around edges and small targets \cite{WTConv}. Third, most methods remain tailored to single-band infrared imagery, and spectral-correlation modeling for hyperspectral data is still insufficient.

Accordingly, a key challenge in infrared hyperspectral colorization is to jointly model spatial structures, frequency-domain textures, and spectral correlations under controllable computational cost. Recently, state space models (SSMs) \cite{Gu_Selective_Scan} offer linear complexity for long-sequence modeling, and Mamba, combining SSMs with selective scanning, provides Transformer-like performance at a lower cost \cite{FreMamba}. However, applying the Mamba architecture to infrared hyperspectral imagery remains nontrivial. The inherently low thermal gradients of infrared data tend to suppress high-frequency details~\cite{ding2025wavelet, TH-Mamba}, posing a fundamental challenge for feature extraction. Furthermore, standard Mamba blocks tend to smooth out local textures during long-sequence state transitions, thereby weakening informative frequency cues. Additionally, the lack of explicit local spatial modeling mechanisms limits the network's capability to effectively represent features amidst scale and structural variations~\cite{DVMSR,SpaMB,MambaFormerSR}.

To address the above issues, we formulate infrared hyperspectral image colorization as a spatial–spectral–frequency coupled generation problem and propose FSCM, a spectral-information-guided conditional GAN framework. Rather than directly extending single-band infrared colorization models to hyperspectral inputs, FSCM introduces a frequency-enhanced spatial–spectral state-space generator to associate high-dimensional infrared spectral responses with spatial structures and natural visible colors. Specifically, the generator is composed of cascaded frequency-cooperative state-space groups (FSGs), each built from several frequency-enhanced spatial–spectral Mamba blocks (FSBs). As the core generation unit, each FSB couples three complementary pathways: a global spatial branch, a local structural branch, and a spectral branch. The global branch employs a visual state space module (VSSM) \cite{VMamba} to capture long-range spatial dependencies with linear complexity, while the spectral branch models inter-band correlations to exploit hyperspectral information. To compensate for the weak textures and degraded boundaries of infrared images, the FEM is embedded into the spatial pathways, where multi-level wavelet decomposition separates structural contours and directional details, and Fourier gating adaptively calibrates global frequency responses. The local branch further introduces the DGM to refine geometry-adaptive local structures and suppress redundant background responses. In this way, FSCM unifies long-range spatial modeling, spectral dependency learning, and frequency-domain detail compensation within the basic generation block, thereby improving structural fidelity and natural color generation. In addition, an online semantic segmentation-guided loss is incorporated to impose semantic-level structural constraints on the generated images, reducing small-object misses and category confusion in complex road scenes.

\begin{figure}[t]
  \centering
  \includegraphics[width=0.85\linewidth]{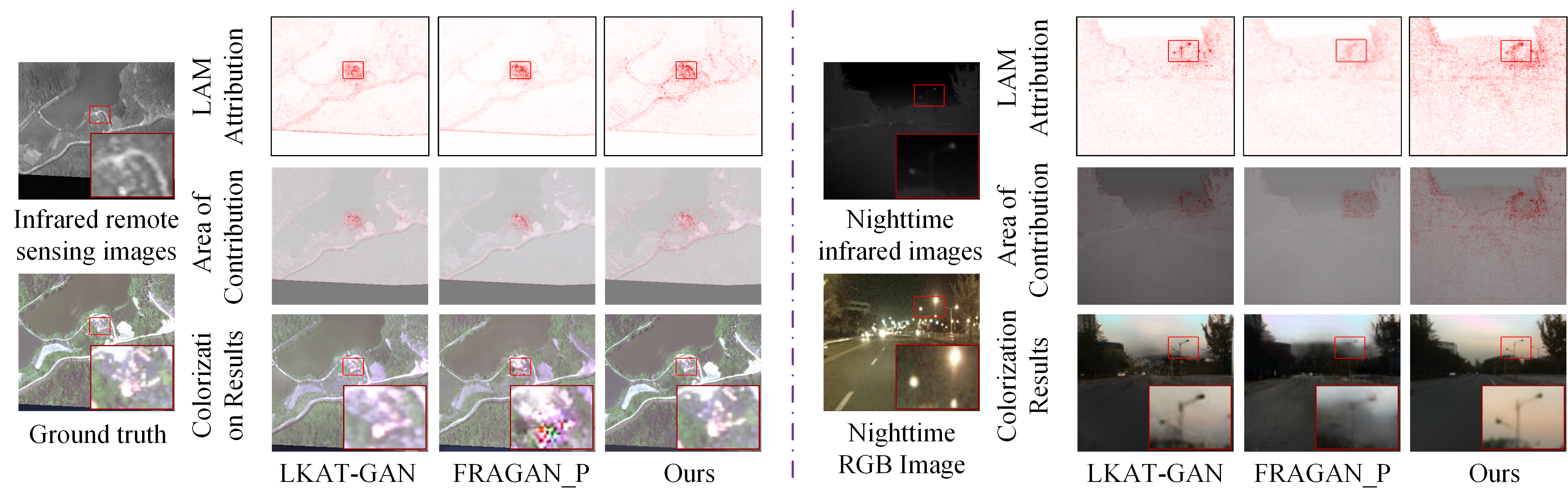}
  \caption{LAM \cite{LAM} visualizations of different networks are presented to reflect the relative contribution of each input pixel to the reconstruction of the patch highlighted by the red box.}
  \label{fig:lam_visualization}
\end{figure}

As illustrated in \autoref{fig:lam_visualization}, local attribution maps (LAM) \cite{LAM} are employed to visualize the colorization results across different methods. By leveraging back-propagated gradients to localize attention regions, LAM characterizes the effective receptive field. Compared with the other two methods, the proposed model demonstrates a broader attention scope, exhibiting stronger suppression of structural distortions and superior color naturalness. These advantages are particularly pronounced in nighttime infrared scenarios.

In summary, the key contributions of this paper are:

(1) A spectral information-guided GAN framework, termed FSCM, is proposed for infrared hyperspectral image colorization. A frequency-enhanced spatial-spectral state-space generator is constructed, in which cascaded FSB units jointly model long-range spatial dependencies and inter-band spectral correlations, thereby supporting structure-aware and spectrally guided color generation.

(2) The FSB unit incorporates detail restoration and local structure refinement to address weak textures and geometric variations in infrared features. Within this unit, FEM restores structural contours and directional high-frequency details through multi-level wavelet decomposition and Fourier gating, while DGM refines geometrically adaptive structures through deformable local sampling and attention sparsification, suppressing redundant background responses and enhancing local structural consistency.

(3) An online semantic segmentation-guided loss is introduced to impose semantic-level structural constraints on the generated images. By supervising segmentation predictions with ground-truth semantic masks, this loss improves semantic consistency, boundary delineation, and small-target retention in complex road scenes.

The remainder of this paper is organized as follows. Section 2 briefly reviews related work on infrared hyperspectral image colorization. Section 3 details the proposed method and the implementation of each module. Section 4 presents qualitative and quantitative experiments to validate the effectiveness of the proposed method. Finally, Section 5 concludes the paper.


\section{Related Work}
\label{sec2}

\subsection{Deep Generative Paradigms for Infrared Colorization}
\label{subsec1}

Due to the imaging mechanism, thermal infrared images are typically low-contrast grayscale signals with weakened fine textures. Early pseudo-color mapping methods relied on hand-crafted priors, such as look-up tables and histogram-based rules, which have limited generalization ability in complex scenes. With the development of deep learning, infrared colorization has become increasingly data-driven, mainly involving CNN-based regression and GAN-based generation methods~\cite{LKATGAN}.

CNN-based methods usually adopt encoder--decoder backbones, such as UNet and ResNet, to learn single-band infrared-to-RGB mapping under pixel-wise and perceptual constraints~\cite{TIR}. These methods can recover plausible colors in simple scenes, but their local receptive fields make it difficult to capture long-range dependencies and global semantic relationships. As a result, blurred edges, missing textures, and unstable colors may appear in traffic and urban scenes. To improve perceptual realism, GAN-based methods, such as pix2pix~\cite{Pix2pix} and its variants~\cite{TICCGAN}, introduce adversarial constraints together with reconstruction losses. Recent methods, including LKAT-GAN~\cite{LKATGAN}, MornGAN~\cite{Morngan}, FRAGAN~\cite{FRAGAN}, and MUGAN~\cite{MUGAN}, further improve multi-scale fusion, attention modeling, and semantic representation, leading to better structural fidelity and color naturalness.

Despite these advances, existing infrared colorization methods still face two main limitations. First, convolution- and attention-based architectures often need to balance global dependency modeling with computational efficiency, which makes it difficult to maintain semantic consistency in complex scenes. Second, most methods are designed for single-band infrared images, where the lack of spectral cues limits material discrimination and may lead to ambiguous colors, blurred boundaries, and structural distortions in fine-grained regions~\cite{MTSIC}. These limitations suggest that richer infrared spectral information may provide useful cues for improving colorization quality.

\subsection{Infrared Hyperspectral Image Colorization}
\label{subsec2}

Infrared hyperspectral imaging records target radiance responses over continuous narrow spectral bands, forming a high-dimensional spectral cube that contains spatial structures, band-wise variations, and material-related information~\cite{HSI_ROAD}. Compared with single-band infrared images, infrared hyperspectral data provide stronger material characterization capability and offer additional cues for reducing semantic ambiguity and improving target discrimination~\cite{BGSR}. However, infrared hyperspectral images still lack natural color representation and often suffer from low texture contrast, weak edge details, and unclear target structures. At present, studies on infrared hyperspectral image colorization remain limited.

TeX Vision is a representative physics-driven method in this direction. Originating from the HADAR framework~\cite{Nature}, it decomposes observed radiance into physical quantities such as temperature, emissivity, and texture through a thermal radiation transfer model, and then generates pseudo-color images in the HSV color space. This method exploits the material characterization capability of infrared hyperspectral data from a physical perspective and provides an interpretable solution for infrared scene perception. However, thermal radiation inversion is inherently ill-posed, because different combinations of physical parameters may correspond to similar observed radiance. In addition, its performance depends on material libraries, environmental parameters, and imaging priors, which limits its adaptability to out-of-library materials, complex backgrounds, and atmospheric interference. The pixel-wise inversion and optimization process also introduce considerable computational cost. Moreover, this physics-driven paradigm is difficult to extend to broader infrared colorization scenarios, especially general infrared hyperspectral scenes with complex material distributions and conventional single-band infrared images, where reliable spectral decomposition and material-library matching are unavailable.

In addition to physics-driven methods, hyperspectral image processing has made substantial progress in band selection, spatial-spectral feature modeling, cross-domain adaptation, image fusion, and super-resolution reconstruction. These studies mainly focus on preserving spectral discriminability, restoring spatial resolution, or integrating cross-modal information, and have provided effective tools for hyperspectral data understanding and reconstruction. However, their primary objective is usually not to establish a generative mapping from infrared hyperspectral radiance responses to natural visible colors. Therefore, they cannot directly address the color representation requirements of infrared hyperspectral image colorization.

These observations indicate that infrared hyperspectral image colorization is different from both conventional hyperspectral classification, fusion, or reconstruction tasks and single-band infrared image colorization. The former mainly emphasizes spectral information preservation and spatial structure restoration, whereas the latter infers colors primarily from single-channel intensity and spatial textures. In contrast, infrared hyperspectral image colorization requires an effective association among high-dimensional spectral responses, infrared spatial structures, and natural color representation. This motivates a data-driven framework that learns semantic and color mappings from multi-band infrared spectral images to visible-light images, providing a flexible solution for complex scenes and broader infrared colorization applications.

\subsection{State-Space Models for Hyperspectral Images}
\label{subsec3}

State-space models have recently provided an efficient solution for long-sequence modeling. Structured state-space models represented by S4 can capture long-range dependencies with linear complexity~\cite{S4}, while Mamba further introduces selective state updating and hardware-friendly scanning. These designs improve sequence feature selection and global dependency modeling while maintaining computational efficiency, making Mamba an attractive choice for high-dimensional visual representation learning.

In computer vision, methods such as Vim~\cite{Vim} and VMamba~\cite{VMamba} extend Mamba to two-dimensional image modeling through bidirectional or multi-directional scanning, enabling effective aggregation of large-range spatial context. Hyperspectral images contain dense spectral bands, strong inter-band correlations, and evident spatial-spectral coupling. These properties are well aligned with the long-sequence modeling capability of Mamba. Accordingly, Mamba-based architectures have been introduced into hyperspectral image classification, fusion, restoration, and super-resolution, where spectral scanning, spatial directional scanning, or joint spatial-spectral modeling is used to enhance band continuity and long-range spatial representation.

Existing hyperspectral Mamba-based methods mainly focus on discriminative representation and image reconstruction~\cite{MFDP,Irsrmamba}, such as improving classification accuracy, enhancing spatial-spectral features, restoring spatial resolution, or preserving spectral consistency. Their modeling objectives are generally confined to the hyperspectral data domain. Infrared hyperspectral image colorization is different because it requires a cross-domain generative transformation from high-dimensional infrared radiance responses to natural RGB images. In this task, spectral variations should be associated not only with spatial structures, but also with semantic regions and plausible visible colors.

For infrared hyperspectral colorization, the long-range modeling capability of Mamba is useful for capturing global spatial structures and inter-band correlations. However, spatial-spectral sequence modeling alone is insufficient, because infrared images usually contain weak high-frequency textures, blurred object boundaries, and background interference. Color generation also requires consistent associations among spectral differences, spatial structures, and semantic regions. Therefore, a colorization-oriented state-space framework should combine spatial-spectral dependency modeling with frequency-domain detail compensation and local structure refinement. This design enables spectral information utilization, boundary preservation, texture recovery, and natural color generation to be improved within a unified generative framework.

\section{Methodology}
\label{sec3}

This section first presents the overall architecture of FSCM. Then, the frequency-cooperative state-space group (FSG) and its core component, the Frequency-Enhanced Spatial-Spectral Mamba Block (FSB), are described in detail. The key modules inside FSB, including the frequency enhancement module (FEM), the multi-domain fusion module (MDFM), and the dual-stream hybrid gating module (DGM), are further explained. Finally, the discriminator and the composite loss functions are described, with emphasis on the online semantic segmentation-guided loss.

\subsection{Overall Architecture}

\begin{figure*}[t]
  \centering
  \includegraphics[width=0.95\linewidth]{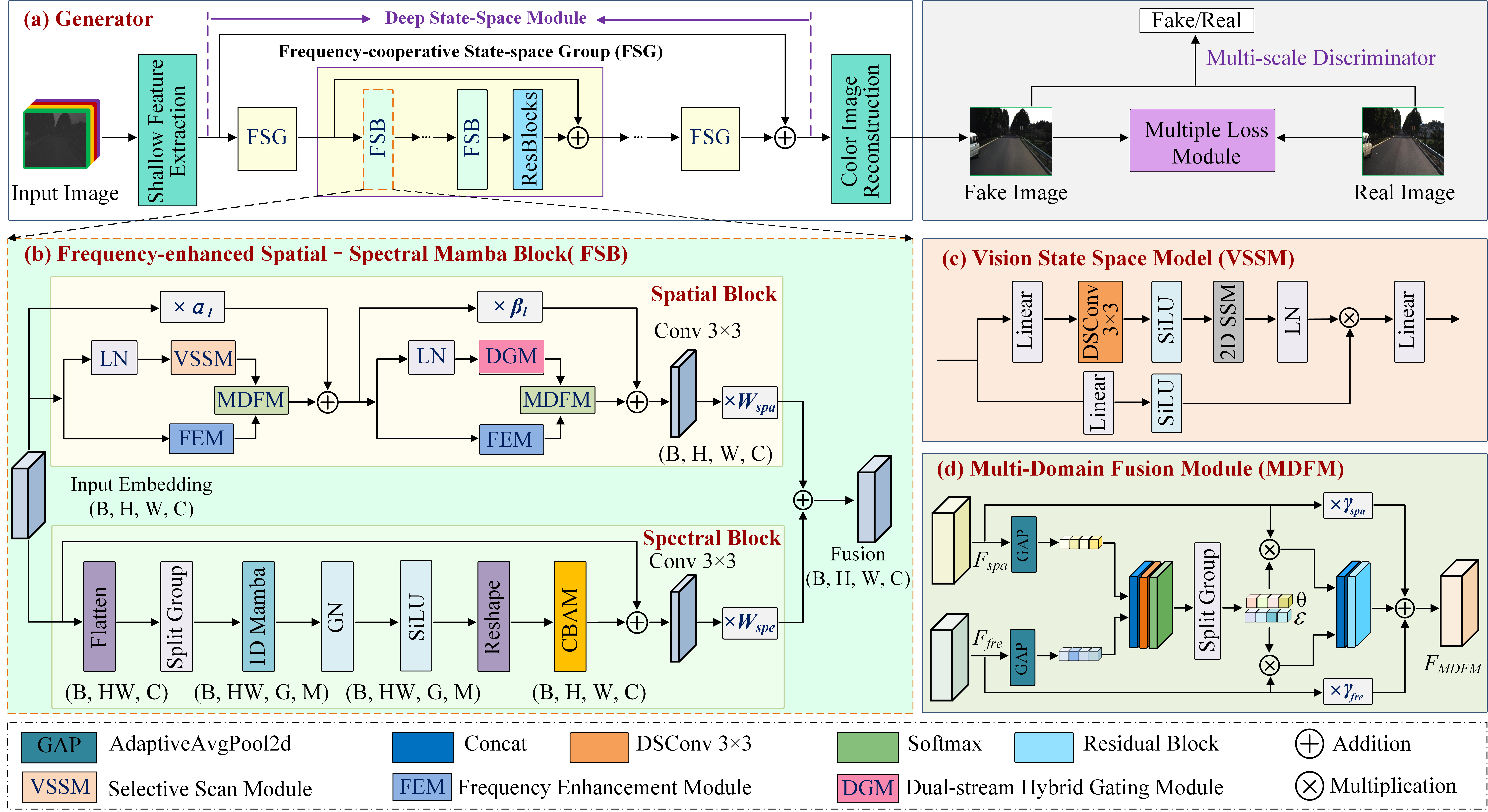}
  \caption{Overview of the proposed FSCM. The generator is built by cascading FSGs, each containing multiple FSBs. Within an FSB, VSSM, DGM, and the Spectral Mamba Branch are integrated, while FEM enhances high-frequency cues and MDFM performs multi-domain feature fusion. $\alpha_l$ and $\beta_l$ denote learnable adaptors for hybrid adaptive integration in the $l$-th FSB.}
  \label{fig:FSCM_overview}
\end{figure*}

As shown in \autoref{fig:FSCM_overview}(a), FSCM follows a GAN framework consisting of a generator, a multi-scale discriminator, and a composite loss module. The generator learns a cross-domain nonlinear mapping from the infrared hyperspectral input $\mathbf{I}_{\mathrm{hsi}} \in \mathbb{R}^{H \times W \times L}$ to the visible prediction $\hat{\mathbf{I}}_{\mathrm{rgb}} \in \mathbb{R}^{H \times W \times 3}$, where $L$ denotes the number of spectral bands. The generator contains three stages: shallow feature extraction, deep state-space refinement, and image reconstruction. Specifically, $\mathbf{I}_{\mathrm{hsi}}$ is first embedded into a unified feature space by convolution, followed by pixel downsampling to reduce the spatial resolution and computational cost while retaining salient information. The shallow feature $\mathbf{F}_0$ is defined as
\begin{equation}
  \mathbf{F}_0 =
  \mathrm{Down}\!\left(
  \mathrm{Conv}\!\left(\mathbf{I}_{\mathrm{hsi}}\right)
  \right),
\end{equation}
where $\mathrm{Conv}(\cdot)$ denotes the convolution operation, and $\mathrm{Down}(\cdot)$ denotes the sub-pixel downsampling operator.

The deep state-space refinement stage is the core of the generator. It consists of $N_G$ cascaded FSGs, where each FSG contains multiple FSBs and residual blocks to progressively refine spatial context, spectral dependencies, and frequency-aware representations. The refined feature is denoted as $\mathbf{F}_{\mathrm{deep}}$.

In the reconstruction stage, a global skip connection fuses $\mathbf{F}_0$ and $\mathbf{F}_{\mathrm{deep}}$ to form a cross-layer information pathway. The fused feature is then upsampled by sub-pixel upsampling $\mathrm{Up}(\cdot)$ and mapped to the visible domain through a reconstruction head $\mathcal{R}(\cdot)$, which consists of a residual group and a $3\times3$ convolution. The generated visible image is obtained as
\begin{equation}
  \hat{\mathbf{I}}_{\mathrm{rgb}}
  =
  \mathcal{R}\!\left(
  \mathrm{Up}\!\left(
  \mathbf{F}_{\mathrm{deep}}+\mathbf{F}_0
  \right)
  \right).
\end{equation}

The generator is optimized with the discriminator and composite losses to improve visual realism and semantic fidelity. The main components are described below.

\subsection{Frequency-Enhanced Spatial-Spectral Mamba Block (FSB)}

As illustrated in \autoref{fig:FSCM_overview}(b), each FSG is composed of multiple FSBs. As the basic generation unit, an FSB contains two spatial paths and one spectral path. The two spatial paths are used for global spatial modeling and local structure refinement, while the spectral path models inter-band dependencies.

Taking $\mathbf{X}^{(l)}_{\mathrm{in}}\in\mathbb{R}^{B\times H\times W\times C}$ as the input of the $l$-th FSB, the global spatial path first applies layer normalization and then processes the feature with VSSM and FEM in parallel. VSSM captures long-range spatial dependencies through two-dimensional selective scanning, while FEM provides frequency-domain cues for high-frequency enhancement. Their outputs are fused by MDFM, scaled by a learnable coefficient $\alpha_l$, and added to the input feature to obtain the global spatial representation $\mathbf{X}_{\mathrm{global}}$.

The local spatial path further refines $\mathbf{X}_{\mathrm{global}}$ by replacing VSSM with DGM. DGM captures geometry-adaptive local structures through deformable convolution and suppresses redundant responses through attention sparsification. FEM is also used in this path to provide local frequency compensation. The outputs of DGM and FEM are fused by MDFM, scaled by a learnable coefficient $\beta_l$, and added to $\mathbf{X}_{\mathrm{global}}$. The resulting feature is then processed by a $3\times3$ convolutional spatial block and weighted by $\mathbf{W}_{\mathrm{spa}}$, producing the spatial feature $\mathbf{X}_{\mathrm{spa}}$.

In parallel, the spectral branch groups the input feature along the channel dimension to form local spectral sequences. A 1D Mamba module is used to capture intra-group spectral correlations and inter-group interactions. After group normalization, SiLU activation, reshaping, and CBAM-based channel recalibration, the spectral feature is processed by a $3\times3$ convolutional spectral block and weighted by $\mathbf{W}_{\mathrm{spe}}$, producing the spectral feature $\mathbf{X}_{\mathrm{spe}}$.

Finally, the weighted spatial and spectral features are fused to obtain the output of the $l$-th FSB:
\begin{equation}
  \mathbf{X}^{(l)}_{\mathrm{out}}
  =
  \mathbf{W}_{\mathrm{spa}}\odot \mathbf{X}_{\mathrm{spa}}
  +
  \mathbf{W}_{\mathrm{spe}}\odot \mathbf{X}_{\mathrm{spe}},
\end{equation}
where $\mathbf{W}_{\mathrm{spa}}$ and $\mathbf{W}_{\mathrm{spe}}$ are learnable channel-wise weights, and $\odot$ denotes element-wise multiplication.

\begin{figure*}[t]
  \centering
  \includegraphics[width=0.85\linewidth]{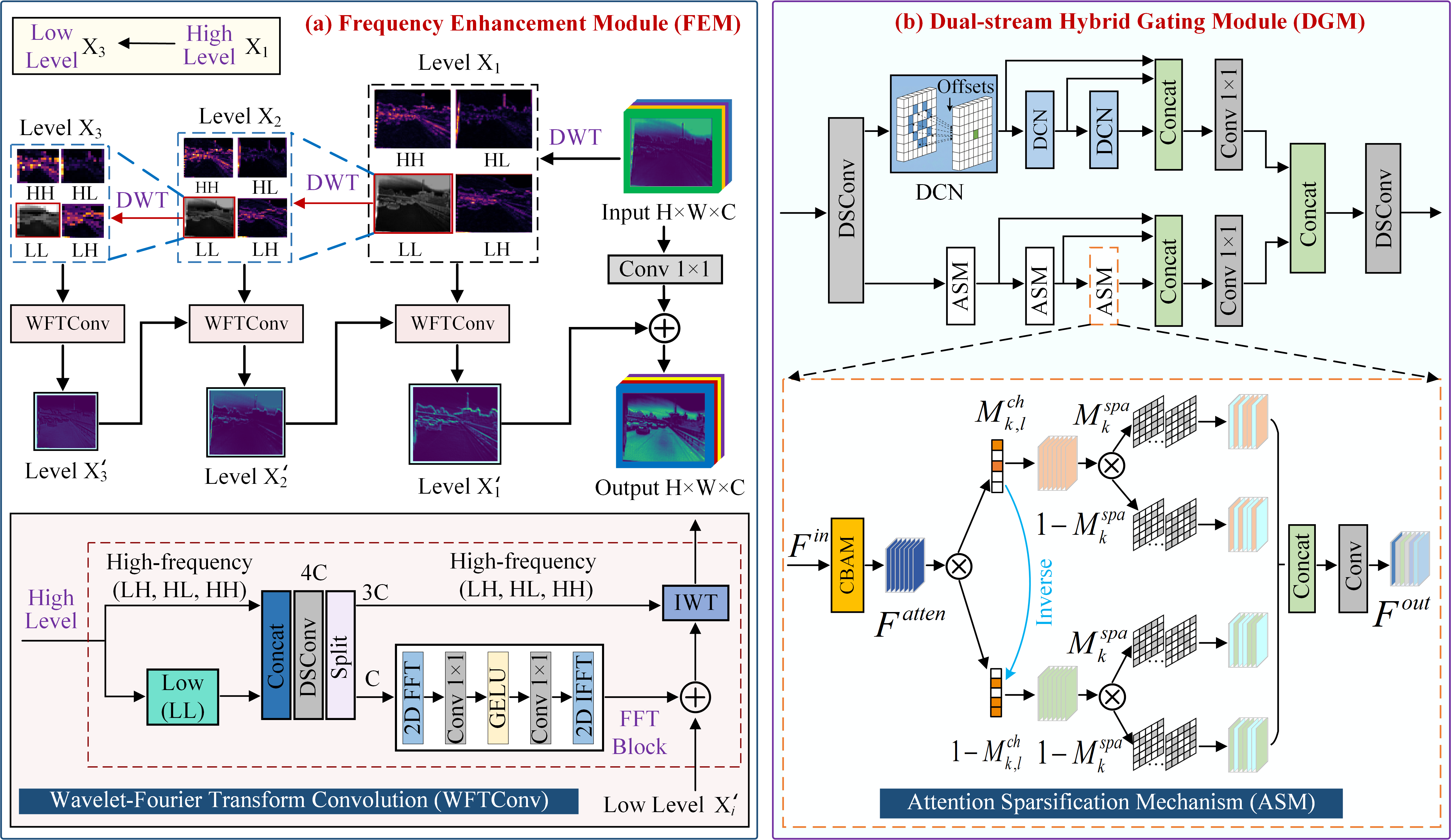}
  \caption{Network architectures of FEM and DGM.}
  \label{fig:FEM_DGM}
\end{figure*}

\textbf{Frequency Enhancement Module (FEM):}
In infrared hyperspectral image colorization, edges, textures, and small-object contours are mainly associated with high-frequency components, whereas regional tones, global structures, and coarse semantic layouts are mainly encoded by low-frequency components. However, infrared images often suffer from weak textures and blurred boundaries due to imaging factors such as thermal diffusion. Spatial-domain convolution or state-space modeling alone is insufficient to explicitly modulate different frequency components. Therefore, FEM, as illustrated in \autoref{fig:FEM_DGM}(a), is introduced to enhance frequency-domain information through Wavelet-Fourier Transform Convolution (WFTConv), which combines multi-level wavelet decomposition, sub-band collaborative enhancement, low-frequency Fourier calibration, and progressive reconstruction.

Given an input feature $\mathbf{X}\in\mathbb{R}^{B\times H\times W\times C}$, a multi-level two-dimensional discrete wavelet transform (DWT) is first recursively applied to the low-frequency sub-band:
\begingroup
\setlength{\jot}{2pt}
\begin{gather}
  \mathbf{X}_{\mathrm{LL}}^{(0)}
  =
  \mathbf{X},
  \\
  \mathbf{X}_{\mathrm{LL}}^{(i)},
  \mathbf{X}_{\mathrm{LH}}^{(i)},
  \mathbf{X}_{\mathrm{HL}}^{(i)},
  \mathbf{X}_{\mathrm{HH}}^{(i)}
  =
  \mathcal{W}\!\left(\mathbf{X}_{\mathrm{LL}}^{(i-1)}\right),
  \quad i=1,\ldots,N .
\end{gather}
\endgroup
where $\mathcal{W}(\cdot)$ denotes the discrete wavelet transform. $\mathbf{X}_{\mathrm{LL}}^{(i)}$ is the low-frequency sub-band at the $i$-th level, which mainly encodes coarse structures and regional tones. $\mathbf{X}_{\mathrm{LH}}^{(i)}$, $\mathbf{X}_{\mathrm{HL}}^{(i)}$, and $\mathbf{X}_{\mathrm{HH}}^{(i)}$ are directional high-frequency sub-bands that capture horizontal, vertical, and diagonal edge and texture information, respectively.

At each decomposition level, the four sub-bands are concatenated along the channel dimension and refined by dilated depthwise separable convolution. The refined feature is then split into enhanced sub-bands, denoted as $\tilde{\mathbf{X}}_{\mathrm{LL}}^{(i)}$, $\tilde{\mathbf{X}}_{\mathrm{LH}}^{(i)}$, $\tilde{\mathbf{X}}_{\mathrm{HL}}^{(i)}$, and $\tilde{\mathbf{X}}_{\mathrm{HH}}^{(i)}$. This operation enables interaction between low-frequency structural information and directional high-frequency details with low computational cost, thereby enhancing boundary and texture responses that are easily attenuated in infrared images.

In this work, a three-level wavelet decomposition is adopted, namely $N=3$. A shallow decomposition with one or two levels mainly focuses on local edge responses, but is less effective in representing larger-scale structural transitions such as road boundaries, building contours, and regional scene layouts. In contrast, deeper decompositions with four or five levels substantially reduce the spatial resolution of the lowest-frequency sub-band, which may suppress small targets and fine boundaries while increasing reconstruction cost. Therefore, three-level decomposition provides a balanced choice among multi-scale representation, detail preservation, and computational efficiency.

Although wavelet decomposition provides local multi-scale and directional frequency representations, the enhanced low-frequency sub-band $\tilde{\mathbf{X}}_{\mathrm{LL}}^{(i)}$ still contains long-range structural correlations and large-scale frequency distributions that are difficult to model sufficiently by local convolution alone. Therefore, Fourier-domain gating is further applied to the enhanced low-frequency sub-band for global frequency calibration:
\begingroup
\setlength{\jot}{2pt}
\begin{gather}
  \mathbf{F}^{(i)}
  =
  \mathcal{F}\!\left(\tilde{\mathbf{X}}_{\mathrm{LL}}^{(i)}\right),
  \\
  \mathbf{Y}^{(i)}
  =
  \mathcal{F}^{-1}\!\left(
  \phi\!\left(\mathbf{F}^{(i)}\mathbf{W}_{1}^{(i)}\right)
  \mathbf{W}_{2}^{(i)}
  \right).
\end{gather}
\endgroup
where $\mathcal{F}(\cdot)$ and $\mathcal{F}^{-1}(\cdot)$ denote the two-dimensional Fourier transform and inverse Fourier transform, respectively, implemented by \texttt{rfft2} and \texttt{irfft2}. $\mathbf{W}_{1}^{(i)}$ and $\mathbf{W}_{2}^{(i)}$ are learnable complex-valued weights, and $\phi(\cdot)$ denotes the GELU activation function. Fourier gating is applied to the low-frequency sub-band because this sub-band mainly controls global tone distribution, regional consistency, and coarse structural responses. Meanwhile, the high-frequency sub-bands are retained in the wavelet domain to preserve spatial localization and directional selectivity.

To avoid excessive perturbation caused by frequency-domain calibration, gated residual modulation is adopted:
\begin{equation}
  \hat{\mathbf{X}}_{\mathrm{LL}}^{(i)}
  =
  \tilde{\mathbf{X}}_{\mathrm{LL}}^{(i)}
  +
  \left(
  2\sigma\!\left(\beta_i\mathbf{Y}^{(i)}\right)-1
  \right)
  \odot
  \mathbf{Y}^{(i)} .
\end{equation}
where $\sigma(\cdot)$ denotes the sigmoid function, $\odot$ denotes element-wise multiplication, and $\beta_i$ is a learnable gating strength parameter initialized to $0.1$. This design adaptively controls the enhancement intensity of frequency responses at different decomposition levels, thereby improving training stability.

After frequency enhancement, the inverse discrete wavelet transform (IWT) is performed in a bottom-up manner to progressively recover the spatial resolution. Let $\mathbf{Z}^{(N+1)}=0$. For $i=N,\ldots,1$, the reconstruction process is formulated as
\begin{equation}
  \mathbf{Z}^{(i-1)}
  =
  \mathcal{W}^{-1}\!\Big(
  \hat{\mathbf{X}}_{\mathrm{LL}}^{(i)}+\mathbf{Z}^{(i+1)},
  \tilde{\mathbf{X}}_{\mathrm{LH}}^{(i)},
  \tilde{\mathbf{X}}_{\mathrm{HL}}^{(i)},
  \tilde{\mathbf{X}}_{\mathrm{HH}}^{(i)}
  \Big),
  \quad i=N,\ldots,1 .
\end{equation}
where $\mathcal{W}^{-1}(\cdot)$ denotes the inverse discrete wavelet transform, and $\mathbf{Z}^{(i-1)}$ is the reconstructed feature after the $i$-th inverse transform. Since Fourier gating is placed between DWT and IWT and acts on the low-frequency sub-band, its global calibration effect can be propagated to higher-resolution features during progressive reconstruction and fused with directional high-frequency details.

Finally, FEM fuses the reconstructed feature with the input feature through a residual shortcut:
\begin{equation}
  \mathbf{Y}_{\mathrm{FEM}}
  =
  \mathrm{Conv}_{1\times1}(\mathbf{X})+\mathbf{Z}^{(0)} .
\end{equation}
where $\mathbf{Y}_{\mathrm{FEM}}$ denotes the output feature of FEM. This residual form preserves the original spatial-spectral representation while compensating for missing boundary and texture information in infrared images.

\textbf{Multi-Domain Fusion Module (MDFM):}
Given the distribution gap between spatial- and frequency-domain features, MDFM is developed for feature alignment and adaptive calibration. It follows a progressive design consisting of global context compression, grouped cross-domain interaction, and residual refinement. Specifically, MDFM first aggregates channel priors from both domains and employs a split-group operation to promote cross-domain channel interaction in the latent space, thereby producing dynamic modulation weights. The calibrated features are then refined by residual blocks to better exploit the complementarity between spatial and frequency-domain information.
\begin{gather}
  \mathbf{F}_{\mathrm{cat}}
  =
  \mathrm{Concat}\Big(
  \mathrm{GAP}(\mathbf{X}_{\mathrm{spa}}),
  \mathrm{GAP}(\mathbf{X}_{\mathrm{fre}})
  \Big),
  \\
  [\boldsymbol{\theta}, \boldsymbol{\varepsilon}]
  =
  \mathcal{S}_{\mathrm{group}}\Big(
  \mathrm{Softmax}\big(
  \mathrm{DSConv}(\mathbf{F}_{\mathrm{cat}})
  \big)
  \Big),
  \\
  \mathbf{F}_{\mathrm{align}}
  =
  \mathcal{R}_{\mathrm{res}}\Big(
  \mathrm{Concat}\big[
  \mathbf{X}_{\mathrm{spa}} \odot \boldsymbol{\theta},
  \mathbf{X}_{\mathrm{fre}} \odot \boldsymbol{\varepsilon}
  \big]
  \Big),
  \\
  \mathbf{F}_{\mathrm{MDFM}}
  =
  \mathbf{F}_{\mathrm{align}}
  +
  \gamma_{\mathrm{spa}}\mathbf{X}_{\mathrm{spa}}
  +
  \gamma_{\mathrm{fre}}\mathbf{X}_{\mathrm{fre}} .
\end{gather}
where $\mathbf{X}_{\mathrm{spa}}, \mathbf{X}_{\mathrm{fre}} \in \mathbb{R}^{H \times W \times C}$ represent the spatial- and frequency-domain inputs, respectively. $\mathrm{DSConv}(\cdot)$ denotes depthwise separable convolution, $\mathrm{GAP}(\cdot)$ denotes global average pooling, $\mathrm{Concat}(\cdot)$ denotes concatenation, and $\mathcal{R}_{\mathrm{res}}(\cdot)$ denotes the residual refinement block. $\mathcal{S}_{\mathrm{group}}(\cdot)$ generates adaptive weights $[\boldsymbol{\theta}, \boldsymbol{\varepsilon}]$, and $\gamma_{\mathrm{spa}}$ and $\gamma_{\mathrm{fre}}$ are set to $0.1$ to preserve the original multi-domain information through residual connections.

\textbf{Dual-Stream Hybrid Gating Module (DGM):}
As depicted in \autoref{fig:FEM_DGM}(b), the Dual-Stream Hybrid Gating Module (DGM) is designed to refine local geometric structures and reduce redundant background responses in infrared features. In infrared images, target contours are often blurred due to thermal diffusion, and local structures may exhibit scale changes and geometric irregularities. Standard convolutions with fixed sampling positions are therefore less effective in capturing such boundary variations. To address this issue, the input feature is first projected by a depthwise separable convolution (DSConv) and then fed into two complementary branches: a deformable convolution (DCN)~\cite{DCN} branch for geometry-adaptive local modeling and an attention-sparsification branch for response selection and redundancy suppression.

Let $\mathbf{X}_{\mathrm{proj}}=\mathrm{DSConv}(\mathbf{X}_{\mathrm{global}})$ denote the projected local feature. In the DCN branch, adaptive offsets are learned to dynamically adjust the sampling grid, enabling the network to better capture irregular object boundaries and local geometric variations. A cascade-dense fusion structure is further used to aggregate multi-stage geometric features:
\begin{gather}
  \mathbf{D}_i
  =
  \Phi^{(i)}_{\mathrm{dcn}}\!\left(\mathbf{D}_{i-1}\right),
  \quad i=1,2,3, \\
  \mathbf{H}_{\mathrm{DCN}}
  =
  \mathrm{Conv}_{1\times1}\!\left(
  \mathrm{Concat}\!\left[
  \mathbf{D}_1,\mathbf{D}_2,\mathbf{D}_3
  \right]
  \right),
\end{gather}
where $\mathbf{D}_0=\mathbf{X}_{\mathrm{proj}}$. $\Phi^{(i)}_{\mathrm{dcn}}(\cdot)$ denotes the $i$-th deformable convolution module, including offset prediction and feature sampling. $\mathbf{H}_{\mathrm{DCN}}$ denotes the extracted geometric feature.

The attention-sparsification branch is based on the Attention Sparsification Mechanism (ASM). It aims to emphasize informative target structures while retaining useful contextual information that may be suppressed by a single attention mask. Specifically, $\mathbf{X}_{\mathrm{proj}}$ is first processed by the Convolutional Block Attention Module (CBAM)~\cite{CBAM} to obtain an attention-enhanced feature $\mathbf{F}_{\mathrm{atten}}$. Then, a channel mask $\mathbf{M}^{\mathrm{ch}}$ and a spatial mask $\mathbf{M}^{\mathrm{spa}}$ are generated to describe channel-wise importance and pixel-wise saliency, respectively:
\begin{gather}
  \mathbf{M}^{\mathrm{ch}}
  =
  \mathrm{Sigmoid}\left(
  \mathrm{Conv}_{1\times1}\left(
  \mathrm{GAP}\left(\mathbf{F}_{\mathrm{atten}}\right)
  \right)
  \right), \\
  \mathbf{M}^{\mathrm{spa}}
  =
  \mathrm{Sigmoid}\left(
  \mathrm{Conv}_{1\times1}\left(\mathbf{F}_{\mathrm{atten}}\right)
  \right).
\end{gather}

Based on these two masks, ASM decomposes the feature into four complementary subspaces, corresponding to different combinations of channel saliency, spatial saliency, and their complementary responses. This decomposition allows the module to enhance high-response target regions, preserve contextual information, and suppress redundant background activations. The initial sparsified feature is formulated as
\begin{equation}
\label{eq:Fout0}
\begin{aligned}
\mathbf{F}^{\mathrm{out}}_0
=
\mathrm{Conv}_{1\times1}\Big(
\mathrm{Concat}\big[
&\mathbf{F}_{\mathrm{atten}}\odot\mathbf{M}^{\mathrm{ch}}\odot\mathbf{M}^{\mathrm{spa}}, \\
&\mathbf{F}_{\mathrm{atten}}\odot\mathbf{M}^{\mathrm{ch}}\odot(1-\mathbf{M}^{\mathrm{spa}}), \\
&\mathbf{F}_{\mathrm{atten}}\odot(1-\mathbf{M}^{\mathrm{ch}})\odot\mathbf{M}^{\mathrm{spa}}, \\
&\mathbf{F}_{\mathrm{atten}}\odot(1-\mathbf{M}^{\mathrm{ch}})\odot(1-\mathbf{M}^{\mathrm{spa}})
\big]\Big).
\end{aligned}
\end{equation}

Finally, an iterative sparsification strategy $\Phi^{(i)}_{\mathrm{sparse}}$ is adopted to progressively refine the selected responses:
\begin{align}
  \mathbf{F}^{\mathrm{out}}_i
  &=
  \Phi^{(i)}_{\mathrm{sparse}}\left(\mathbf{F}^{\mathrm{out}}_{i-1}\right),
  \quad i=1,2,3, \\
  \mathbf{H}_{\mathrm{sparse}}
  &=
  \mathrm{Conv}_{1\times1}\left(
  \mathrm{Concat}\left[
  \mathbf{F}^{\mathrm{out}}_1,
  \mathbf{F}^{\mathrm{out}}_2,
  \mathbf{F}^{\mathrm{out}}_3
  \right]
  \right).
\end{align}
where $\mathbf{H}_{\mathrm{sparse}}$ denotes the output of the attention-sparsification branch. The geometric feature from the DCN branch and the sparsified feature from the ASM branch are then fused to obtain the DGM output:
\begin{equation}
  \mathbf{H}_{\mathrm{out}}
  =
  \mathrm{DSConv}\left(
  \mathrm{Concat}\left[
  \mathbf{H}_{\mathrm{DCN}},
  \mathbf{H}_{\mathrm{sparse}}
  \right]
  \right),
\end{equation}
where $\mathbf{H}_{\mathrm{out}}$ denotes the output feature of DGM.

\subsection{Discriminator}

To generate sharp textures while preserving global semantics and color distribution, we integrate a PatchGAN~\cite{Pix2pix} discriminator with the statistics-based SPatchGAN~\cite{Spatchgan}. PatchGAN is fully convolutional and outputs an $N\times N$ real/fake score map, focusing on high-frequency details and suppressing blurry textures and local artifacts. However, its limited long-range modeling can cause global color shifts or geometric inconsistencies. To mitigate this, SPatchGAN adds a global statistical branch that summarizes multi-scale features (e.g., mean and variance) and predicts discrimination scores via independent MLPs. The two branches are complementary: PatchGAN enforces local realism, while SPatchGAN promotes global statistical consistency. Jointly optimizing the local adversarial loss and the global statistical loss encourages both fine textures and alignment with the real images' global statistical distribution.

\subsection{Loss Function}

Although FSCM can model spatial-spectral dependencies through the proposed generation blocks, infrared-to-visible colorization remains an ill-posed cross-domain translation problem. If only pixel-wise reconstruction loss is used, the generated results may suffer from semantic ambiguity, structural inconsistency, and color distortion. To address this issue, the generator is optimized with a composite loss that combines conditional adversarial supervision, content reconstruction constraints, and online semantic segmentation-guided supervision. The total loss is formulated as
\begin{equation}
  \mathcal{L}_{\mathrm{total}}
  =
  \lambda_{\mathrm{cGAN}}\mathcal{L}_{\mathrm{cGAN}}
  +
  \lambda_{\mathrm{content}}\mathcal{L}_{\mathrm{content}}
  +
  \lambda_{\mathrm{seg}}\mathcal{L}_{\mathrm{seg}},
\end{equation}
where $\mathcal{L}_{\mathrm{cGAN}}$, $\mathcal{L}_{\mathrm{content}}$, and $\mathcal{L}_{\mathrm{seg}}$ denote the conditional adversarial loss, content reconstruction loss, and semantic segmentation-guided loss, respectively. $\lambda_{\mathrm{cGAN}}$ and $\lambda_{\mathrm{content}}$ are set to $0.5$ and $1$, respectively. To avoid unstable semantic constraints in the early stage of color generation, $\lambda_{\mathrm{seg}}$ is initialized as $0$ and increased to $0.5$ from the 50th epoch.

\textbf{Conditional adversarial loss:}
To improve the realism of generated visible images, a conditional GAN loss~\cite{Pix2pix} is used with the infrared hyperspectral image and the corresponding visible image as paired inputs. The conditional adversarial loss is defined as
\begin{equation}
  \begin{aligned}
  \mathcal{L}_{\mathrm{cGAN}}(G,D)
  =
  &\mathbb{E}_{\mathbf{I}_{\mathrm{hsi}},\mathbf{I}_{\mathrm{GT}}}
  \left[
  \log D(\mathbf{I}_{\mathrm{hsi}},\mathbf{I}_{\mathrm{GT}})
  \right] \\
  &+
  \mathbb{E}_{\mathbf{I}_{\mathrm{hsi}}}
  \left[
  \log
  \left(
  1-D(\mathbf{I}_{\mathrm{hsi}},\hat{\mathbf{I}}_{\mathrm{rgb}})
  \right)
  \right],
  \end{aligned}
\end{equation}
where $\mathbf{I}_{\mathrm{hsi}}\in\mathbb{R}^{H\times W\times L}$ denotes the infrared hyperspectral image, $\mathbf{I}_{\mathrm{GT}}\in\mathbb{R}^{H\times W\times 3}$ denotes the ground-truth visible image, and $\hat{\mathbf{I}}_{\mathrm{rgb}}=G(\mathbf{I}_{\mathrm{hsi}})$ denotes the generated visible image.

\textbf{Content loss:}
Following Kuang \emph{et al.}~\cite{TICCGAN}, the content loss is constructed by combining pixel loss $\mathcal{L}_{\mathrm{pix}}$, edge loss $\mathcal{L}_{\mathrm{edge}}$~\cite{Ledge}, frequency-domain loss $\mathcal{L}_{\mathrm{fft}}$~\cite{Lfft}, perceptual loss $\mathcal{L}_{\mathrm{per}}$, structural similarity loss $\mathcal{L}_{\mathrm{ssim}}$, and total-variation loss $\mathcal{L}_{\mathrm{tv}}$~\cite{TICCGAN}. These terms jointly constrain pixel fidelity, boundary preservation, frequency consistency, perceptual similarity, structural consistency, and local smoothness. The content loss is defined as
\begin{equation}
  \begin{aligned}
  \mathcal{L}_{\mathrm{content}}
  =
  &\lambda_{\mathrm{pix}}\mathcal{L}_{\mathrm{pix}}
  +
  \lambda_{\mathrm{edge}}\mathcal{L}_{\mathrm{edge}}
  +
  \lambda_{\mathrm{per}}\mathcal{L}_{\mathrm{per}}  \\
  &+
  \lambda_{\mathrm{ssim}}\mathcal{L}_{\mathrm{ssim}}
  +
  \lambda_{\mathrm{fft}}\mathcal{L}_{\mathrm{fft}}
  +
  \lambda_{\mathrm{tv}}\mathcal{L}_{\mathrm{tv}},
  \end{aligned}
\end{equation}
where $\lambda_{\mathrm{pix}}=10$, $\lambda_{\mathrm{per}}=10$, and $\lambda_{\mathrm{edge}}=\lambda_{\mathrm{tv}}=\lambda_{\mathrm{ssim}}=\lambda_{\mathrm{fft}}=1$ are used to balance different loss terms. This formulation provides both spatial-domain and frequency-domain constraints, thereby improving the structural fidelity and visual quality of infrared colorization.

\begin{figure*}[t]
  \centering
  \includegraphics[width=0.92\linewidth]{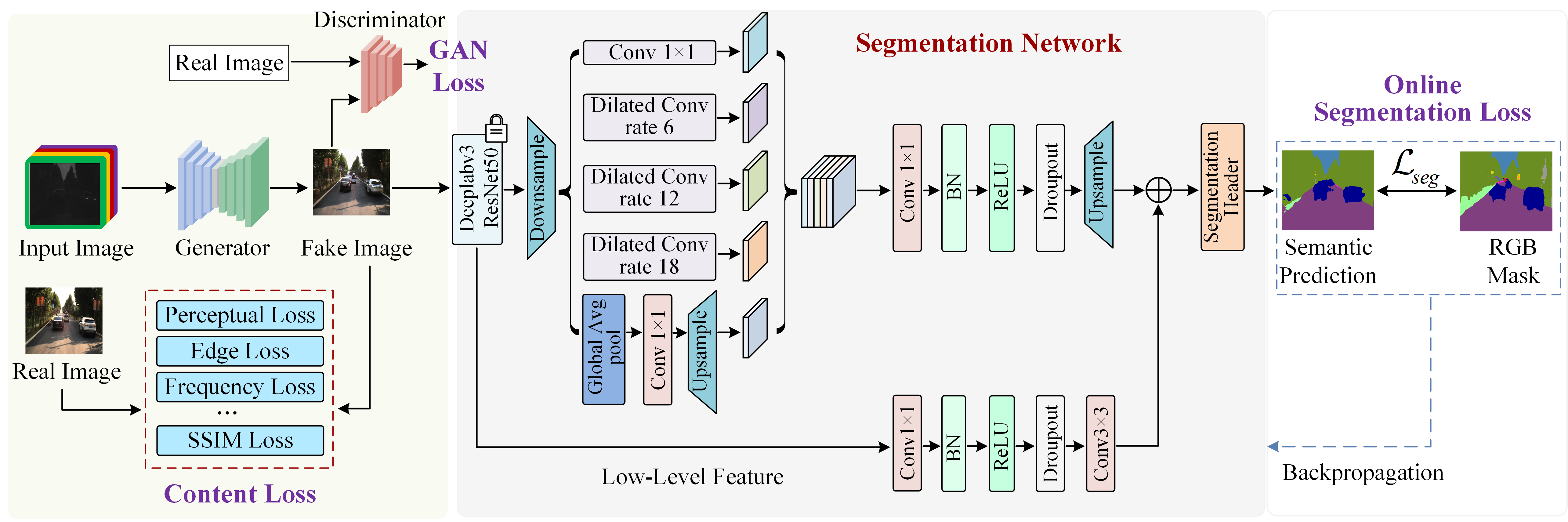}
  \caption{Composite loss module.}
  \label{fig:multiple_loss}
\end{figure*}

\textbf{Online semantic segmentation-guided loss:}
For road-scene infrared hyperspectral image colorization, an online semantic segmentation-guided loss is further used to improve semantic integrity and reduce small-object omissions and boundary blur, as illustrated in \autoref{fig:multiple_loss}. Different from pixel-level reconstruction losses and adversarial loss, this loss provides semantic-level structural supervision for the generated RGB image. It encourages the generated image to preserve object boundaries and category-consistent regions.

A semantic segmentation network $S$ is used as an auxiliary supervisory network. As shown in \autoref{fig:multiple_loss}, $S$ is built on DeepLabv3 \cite{DeepLabv3Plus} with a ResNet-50 \cite{He2016ResNet} backbone. The backbone extracts low-level and high-level features, while the dilated convolution branches capture multi-scale contextual information. The extracted features are fused and upsampled to produce per-pixel semantic class probabilities.

The segmentation network $S$ is first pre-trained on real RGB images with pixel-level semantic annotations $Y_{\mathrm{gt}}$ using the cross-entropy loss. During colorization training, the parameters of $S$ are fixed, and only the gradients with respect to the generated image are backpropagated to update the generator. For a generated image $G(\mathbf{I}_{\mathrm{hsi}})$, the segmentation prediction $S(G(\mathbf{I}_{\mathrm{hsi}}))$ is compared with the corresponding semantic label $Y_{\mathrm{gt}}$. In this way, the generator is encouraged to produce colorized images that are not only visually plausible but also structurally and semantically consistent with the target scene. This constraint is particularly useful for road scenes, where small objects, lane boundaries, vehicles, and object contours are easily blurred or missed during colorization.

The segmentation-guided loss is defined as
\begin{equation}
  \mathcal{L}_{\mathrm{seg}}
  =
  -\frac{1}{HW}
  \sum_{i=1}^{HW}
  \sum_{c=1}^{C}
  Y_{\mathrm{gt}}^{(i,c)}
  \log
  \left(
  S\left(G\left(\mathbf{I}_{\mathrm{hsi}}\right)\right)^{(i,c)}
  +\epsilon
  \right),
\end{equation}
where $C$ denotes the number of semantic classes, $i$ indexes pixels, and $c$ indexes classes. $Y_{\mathrm{gt}}^{(i,c)}$ denotes the one-hot semantic label of class $c$ at pixel $i$, $G(\cdot)$ denotes the generator, $S(\cdot)$ denotes the fixed segmentation network, and $\epsilon$ is a small constant for numerical stability. Note that targets in remote-sensing hyperspectral images are often small, whereas most segmentation models are trained on road-scene datasets; therefore, we do not use segmentation loss for remote sensing scenes to avoid performance degradation caused by unreliable segmentation.

\section{Experiments and Results}

This section presents the infrared hyperspectral datasets, evaluation metrics, and implementation details. The proposed method is compared with representative approaches, and ablation studies are conducted to quantify the contributions of individual modules and key loss terms. Owing to the limited availability of infrared hyperspectral data, systematic investigations remain scarce. Experiments are performed on three infrared hyperspectral datasets, namely HADAR~\cite{Nature}, HSI ROAD~\cite{HSI_ROAD}, and IHSR, with additional evaluations on the KAIST multispectral thermal infrared dataset~\cite{Hwang2015KAIST} to assess robustness under complex scenarios with reduced spectral dimensionality.

\subsection{Datasets}

(a) {HADAR:} A synthetic long-wave infrared (LWIR) dataset with 10 sequences depicting complex nighttime driving scenes. Each frame is a thermal radiance cube of size \(1080 \times 1920 \times 54\) (height \(\times\) width \(\times\) bands).

(b) {HSI ROAD:} A real-world road segmentation dataset with 3799  RGB–near-infrared (NIR) image pairs. It includes 28 NIR bands and covers various road materials and environments, offering spectral richness and scene diversity.

(c) {IHSR:} A real-world LWIR hyperspectral dataset is collected, Zhejiang Province, China. It covers mountains, urban areas, rivers, parks, highways, and farmland, with a spatial resolution of 1.0~m and a spectral range of \(8.0\text{--}11.3~\mu\mathrm{m}\), spanning 110 contiguous bands.

\subsection{Implementation Details}

We evaluate colorization on HADAR in nighttime road and urban traffic scenes. Due to GPU memory limits, hyperspectral cubes are cropped into \(460 \times 512 \times 54\) patches with a 64-pixel overlap for training. Testing is done at the original resolution without cropping or geometric augmentation. For HSI ROAD, images are resized from \(192 \times 384\) to \(512 \times 450\). The dataset has paired RGB and hyperspectral images, split into 3686 training pairs and 113 testing pairs. For IHSR, pixel-level registration is performed in ENVI. We randomly select 20 non-overlapping regions for testing and use the remainder for training. Sliding-window sampling with stride 24 is applied for augmentation, resulting in 2386 training pairs and 20 testing pairs. Each sample is a \(354 \times 354 \times 110\) hyperspectral patch with its aligned RGB image.

During training, images are randomly cropped to $256 \times 256$. Models are trained using Adam with a batch size of 1 for 200 epochs, starting from a learning rate of $1.2 \times 10^{-4}$, which is kept constant for the first 50 epochs and then linearly decayed to zero. All experiments are conducted on Ubuntu~18.04 with PyTorch~2.0.1 using an NVIDIA RTX~3090 GPU.

\subsection{Evaluation Metrics}

To evaluate both physical fidelity and perceptual quality, we use full-reference and no-reference metrics for infrared hyperspectral colorization. Specifically, PSNR, SSIM \cite{SSIM}, NIQE \cite{NIQE}, and UIQI \cite{UIQI} are reported. PSNR and SSIM measure content and structural consistency with ground-truth visible images (higher is better, with SSIM having an ideal value of 1). NIQE evaluates perceptual naturalness without reference (lower is better), while UIQI quantifies agreement in luminance, contrast, and structure, with an ideal value of 1.

\subsection{Comparisons With State-of-the-Art Methods}

Comparisons are conducted with nine representative infrared colorization and image-to-image translation methods, including pix2pix~\cite{Pix2pix}, TICC-GAN~\cite{TICCGAN}, ToDayGAN~\cite{TodayGAN}, FRAGAN\_P~\cite{FRAGAN}, LKAT-GAN~\cite{LKATGAN}, DDGAN~\cite{DDGAN}, MUGAN~\cite{MTSIC}, MornGAN~\cite{Morngan}, and VOS~\cite{VOS}. These methods were originally designed mainly for single-band infrared image colorization or RGB image-to-image translation. Since dedicated methods for infrared hyperspectral image colorization are still limited, they are selected as representative baselines to evaluate color restoration, texture reconstruction, and structural preservation.

For a fair comparison, all methods are retrained on each dataset using the same training and testing splits as the proposed method. The input layers of the compared methods are adapted to match the spectral channel number of each dataset, while the remaining network architectures, training strategies, and publicly released or recommended hyperparameters are kept consistent with their original settings. Therefore, all methods are evaluated under the same input data setting and experimental protocol. Notably, we adopt the supervised FRAGAN\_P, which also serves as the backbone for VOS.

\begin{figure*}[htbp]
  \centering
  \includegraphics[width=1\linewidth]{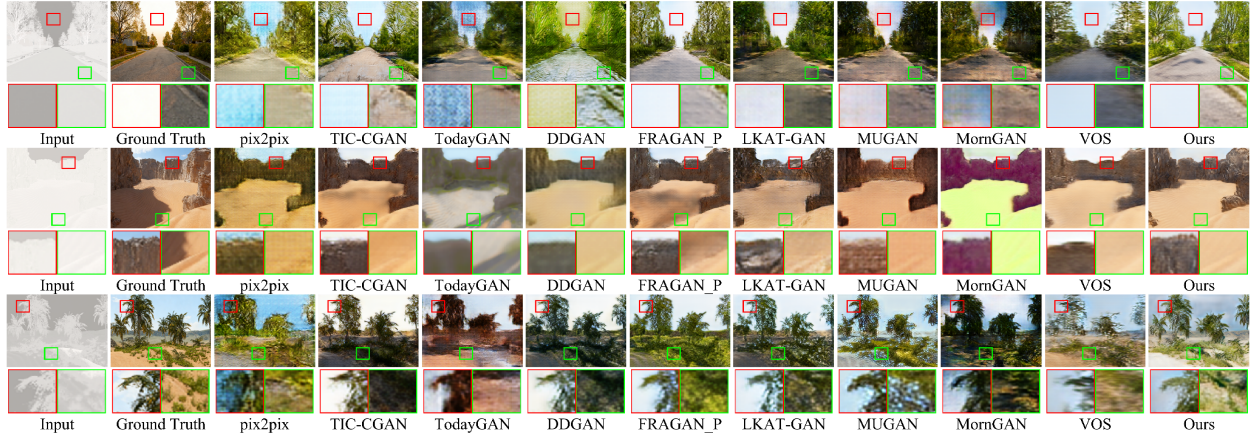}
  \caption{Visual comparison of sample images from the HADAR thermal infrared hyperspectral dataset.}
  \label{fig:hadar_visual}
\end{figure*}

\subsubsection{\textbf{Qualitative Analysis}}

\autoref{fig:hadar_visual} presents the visual comparison on the HADAR thermal infrared hyperspectral dataset. The input LWIR images contain limited texture and weak contrast, making reliable color recovery challenging. pix2pix~\cite{Pix2pix} and TICC-GAN~\cite{TICCGAN} tend to generate desaturated results with blurred local structures, and the separation between roads, vegetation, and buildings is not sufficiently clear. ToDayGAN~\cite{TodayGAN} and DDGAN~\cite{DDGAN} produce stronger colors, but visible color shifts and artifacts appear in transition regions. Other GAN-based methods, such as FRAGAN\_P~\cite{FRAGAN}, LKAT-GAN~\cite{LKATGAN}, MUGAN~\cite{MUGAN}, and MornGAN~\cite{Morngan}, improve the overall appearance to some extent, but still suffer from local texture loss or unstable tones. In comparison, the proposed method generates more natural global colors and better preserves boundaries and local details in the highlighted regions, showing improved structural fidelity under low-quality infrared imaging conditions.

\begin{figure*}[htbp]
  \centering
  \includegraphics[width=1\linewidth]{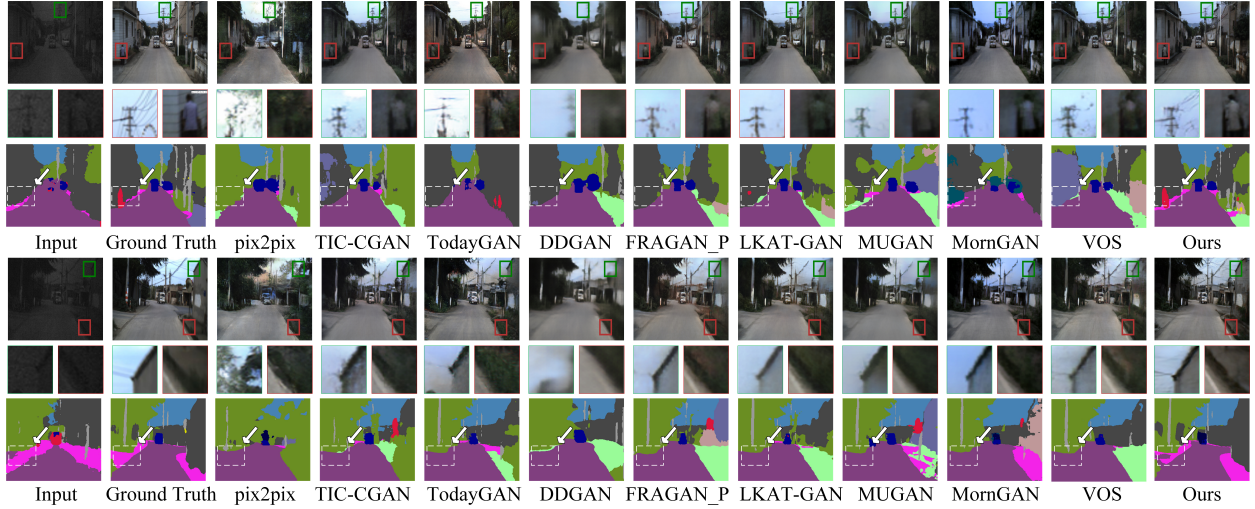}
  \caption{Visual comparison on the HSI ROAD dataset, including colorization results, enlarged local regions, and the corresponding semantic segmentation maps. The regions highlighted by white dotted boxes in the segmentation maps are of particular interest.}
  \label{fig:hsi_road_comparison}
\end{figure*}

\autoref{fig:hsi_road_comparison} shows the visual results on the HSI ROAD dataset, including colorization results, enlarged local regions, and the corresponding segmentation maps. For the colorization results, several baseline methods produce blurred distant structures, unstable brightness, or color inconsistency, especially around thin objects such as poles, vehicles, and roadside boundaries. These local degradations also affect the segmentation results, where object regions may become fragmented or poorly aligned with boundaries. By contrast, the proposed method produces clearer object contours and more consistent road-scene colors. The enlarged regions show that thin structures and local boundaries are better retained, and the segmentation maps are more continuous and better aligned with the scene layout. This indicates that the proposed method can provide more discriminative structural cues for downstream semantic understanding in road scenes.

\begin{figure*}[htbp]
  \centering
  \includegraphics[width=1\linewidth]{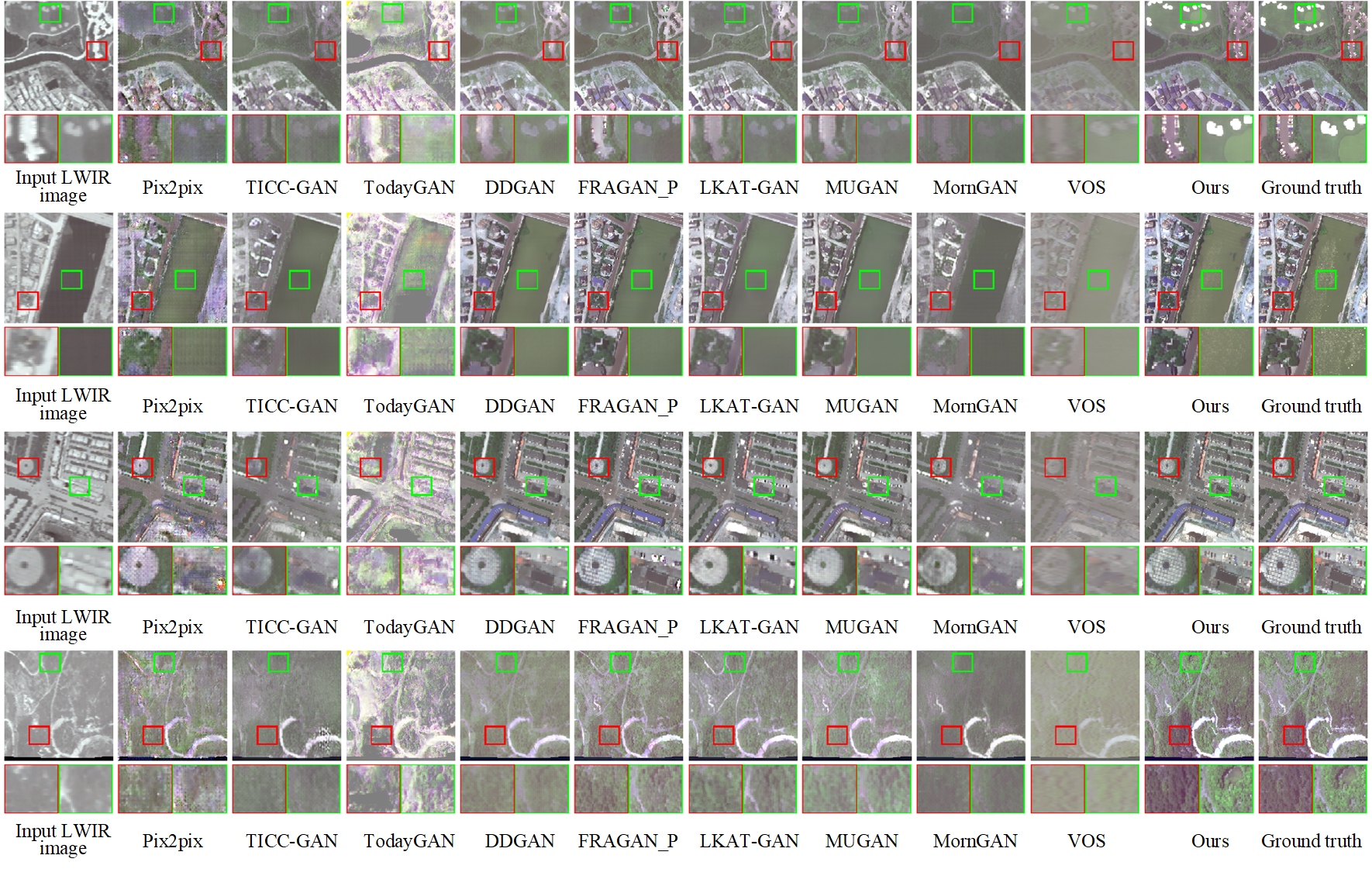}
  \caption{Visual comparison of sample images from the IHSR long-wave infrared hyperspectral remote-sensing dataset.}
  \label{fig:IHSR_visual}
\end{figure*}

\autoref{fig:IHSR_visual} compares different methods on the IHSR long-wave infrared hyperspectral remote-sensing dataset. This dataset contains complex natural scenes and high-dimensional spectral responses, where color hallucination, texture distortion, and boundary ambiguity are common problems. pix2pix, TICC-GAN, and VOS often generate low-contrast or over-smoothed results, while ToDayGAN, DDGAN, and MornGAN may introduce obvious color bias or block artifacts in vegetation, road, and water-land transition regions. Although some methods can recover partial scene colors, fine textures and object boundaries are still not well preserved. The proposed method achieves more stable color tones and clearer structural transitions, especially in the highlighted local regions. These results further demonstrate the benefit of jointly exploiting spatial, spectral, and frequency information for infrared hyperspectral image colorization.

\subsubsection{\textbf{Experimental Extension on the Nighttime KAIST Dataset}}

The KAIST dataset~\cite{KAIST} contains challenging nighttime traffic and campus scenes with complex road layouts, buildings, vehicles, pedestrians, and other fine-grained objects. Since paired daytime color ground truth is not available for this dataset, semantic segmentation is used as an auxiliary evaluation to examine whether colorization can improve the semantic interpretability of thermal infrared images.

\begin{figure*}[htbp]
  \centering
  \includegraphics[width=0.95\linewidth]{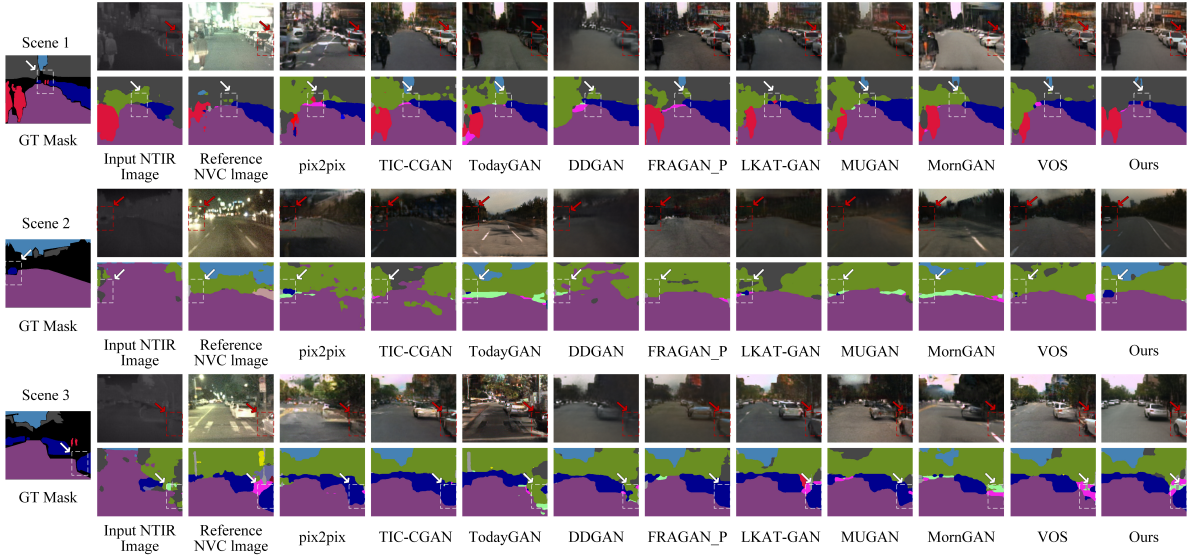}
  \caption{Visual comparison of colorization (top row) and segmentation (bottom row) results on the KAIST dataset. The regions enclosed in white dotted boxes are of particular interest.}
  \label{fig:kaist_visual}
\end{figure*}

All colorization methods improve the mIoU over the raw NTIR input, which suggests that cross-modal colorization can partially reduce the modality gap and enhance semantic readability. Conventional methods such as pix2pix, ToDayGAN, and DDGAN achieve relatively limited improvements, with mIoU values around 25-27. More advanced methods, including TICC-GAN, FRAGAN\_P, LKAT-GAN, MUGAN, MornGAN, and VOS, further improve the performance, but their segmentation results still show fragmented regions, unstable boundaries, or missing small objects in complex scenes.

The qualitative results in \autoref{fig:kaist_visual} are consistent with the quantitative comparison. Baseline methods can improve the visibility of major regions to some extent, but they often produce inaccurate boundaries or fragmented segmentation maps, particularly around distant vehicles, pedestrians, road edges, and building-sky interfaces. In contrast, the proposed method produces more continuous road regions, clearer building-sky separation, and better preserved vehicle and pedestrian regions. This demonstrates that the proposed colorization framework is not only effective for visual reconstruction, but also beneficial for downstream semantic understanding of nighttime thermal infrared scenes.

\subsection{Ablation Study}

Ablation studies are conducted to evaluate the contributions of FSG, FSB, and the key components in FSB, including FEM, multi-domain fusion module (MDFM), and DGM. For a fair comparison, all variants use the same data split and training settings, with only the target component or loss term modified each time. Since HSI ROAD and KAIST contain complex road scenes and small objects, the ablation experiments are mainly performed on these datasets to evaluate both colorization quality and semantic consistency.

\subsubsection{\textbf{Overall Effect of FSG}}

\begin{table}[t]
  \centering
  \caption{Ablation study of key components and the segmentation loss in FSG.}
  \label{tab:ablation_fsg}
  \footnotesize
  \setlength{\tabcolsep}{12pt}
  \renewcommand{\arraystretch}{1.08}
  \begin{tabular}{lccccccc}
    \toprule
    \multirow{2}{*}{\textbf{Variant}}       &
    \multicolumn{3}{c}{\textbf{Components}} &
    \multicolumn{4}{c}{\textbf{Metrics}}                                                                                                                                                                 \\
    \cmidrule(lr){2-4}\cmidrule(lr){5-8}
                                            & \textbf{FSB}            & \textbf{RBs}            & $\mathcal{L}_{\mathrm{seg}}$
                                            & \textbf{PSNR}$\uparrow$ & \textbf{SSIM}$\uparrow$ & \textbf{NIQE}$\downarrow$    & \textbf{UIQI}$\uparrow$                                                 \\
    \midrule
    Baseline                                & $\times$                & $\checkmark$            & $\checkmark$                 & 15.53                   & 0.52          & 4.25          & 0.81          \\
    w/o RBs                                 & $\checkmark$            & $\times$                & $\checkmark$                 & 21.82                   & 0.73          & 3.49          & 0.91          \\
    w/o $\mathcal{L}_{\mathrm{seg}}$        & $\checkmark$            & $\checkmark$            & $\times$                     & 21.96                   & 0.76          & 3.51          & 0.93          \\
    \textbf{Ours}                           & $\checkmark$            & $\checkmark$            & $\checkmark$                 & \textbf{22.52}          & \textbf{0.77} & \textbf{3.32} & \textbf{0.94} \\
    \bottomrule
    \multicolumn{8}{l}{\footnotesize RBs denote residual blocks.}
  \end{tabular}
\end{table}

\begin{figure}[t]
  \centering
  \includegraphics[width=\linewidth]{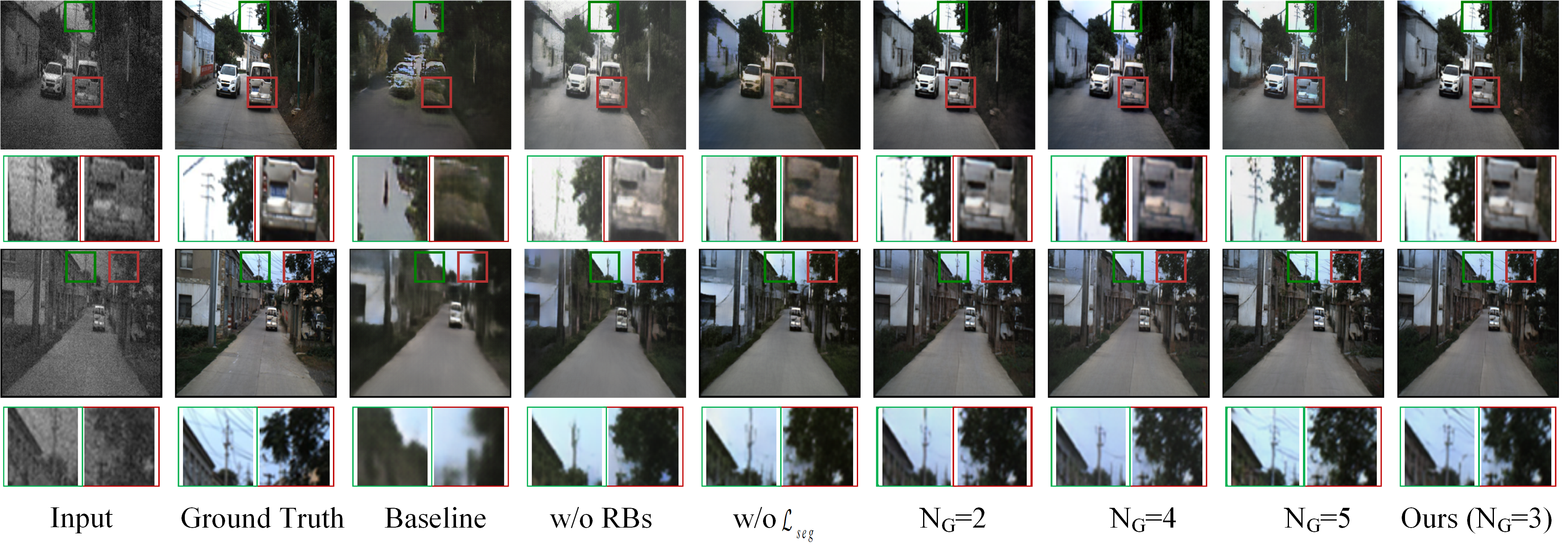}
  \caption{Ablation study of the FSG module. ``w/o'' denotes the removal of the corresponding module.}
  \label{fig:Ablation_FSG}
\end{figure}

As shown in \autoref{tab:ablation_fsg}, the ResBlocks-only baseline obtains only 15.53 dB PSNR and 0.52 SSIM, indicating that purely spatial residual modeling is insufficient to bridge the infrared-visible domain gap. When FSB is retained and RBs are removed, the performance increases substantially to 21.82 dB PSNR and 0.73 SSIM. This demonstrates that FSB plays a major role in improving colorization quality by jointly modeling spatial, spectral, and frequency information.

When the segmentation loss is removed, the model still achieves 21.96 dB PSNR and 0.76 SSIM, confirming that the main reconstruction capability comes from the proposed generator architecture. After $\mathcal{L}_{\mathrm{seg}}$ is added, the full model further improves PSNR to 22.52 dB and reduces NIQE from 3.51 to 3.32. Although the gain is moderate, it improves structural consistency and perceptual quality. The visual results in \autoref{fig:Ablation_FSG} also show that the segmentation-guided loss helps reduce boundary drift and miscoloring in semantically sensitive regions.

\subsubsection{\textbf{Analysis of Key Components within FSB}}

\begin{figure}[t]
\centering
\includegraphics[width=\linewidth]{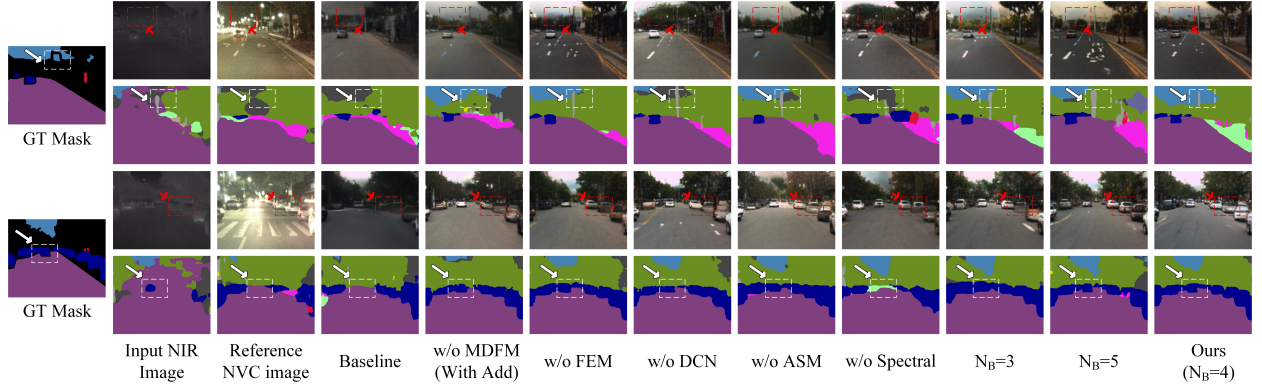}
\caption{Visual ablation analysis of the FSB module. ``w/o'' denotes the removal of the corresponding component.}
\label{fig:FSB_ablation}
\end{figure}

\begin{figure}[t]
\centering
\includegraphics[width=\linewidth]{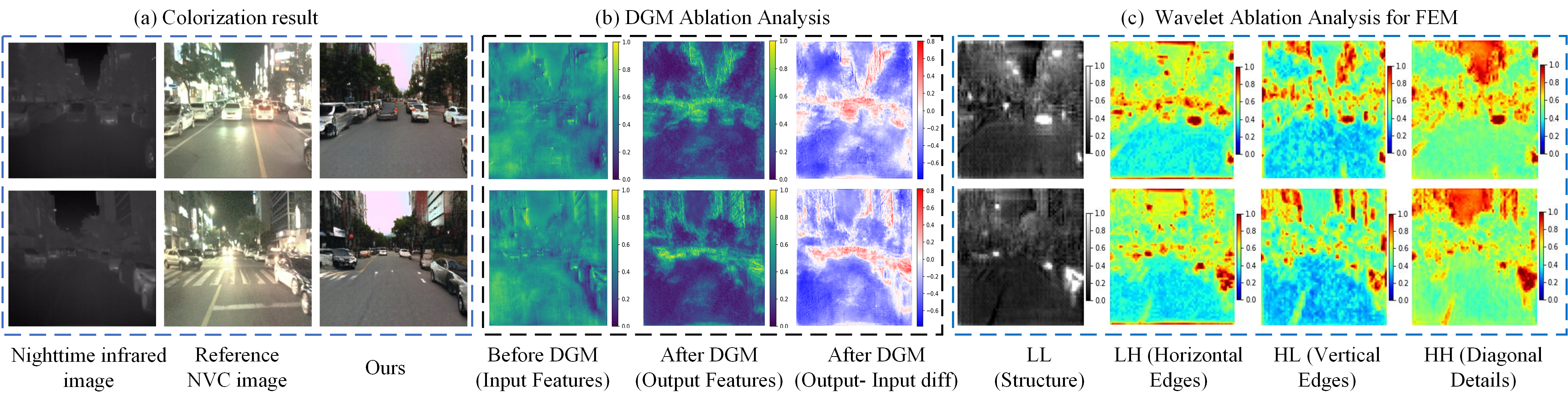}
\caption{Visual analysis of DGM and FEM within FSB.}
\label{fig:Ablation_DGM_FEM}
\end{figure}

The qualitative ablation results of FSB are shown in \autoref{fig:FSB_ablation}. When the whole FSB is removed, the generated results exhibit fragmented semantic regions, blurred object boundaries, and weakened local structures. This indicates that simple spatial residual modeling is insufficient for recovering reliable semantic structures after infrared-to-visible colorization. In contrast, the complete FSB produces more continuous semantic masks and clearer structural transitions, suggesting that the joint use of spatial, spectral, and frequency-aware representations is important for infrared hyperspectral image colorization.

Within FSB, removing MDFM weakens the coordination between spatial-domain and frequency-domain features, leading to less stable structural details. Removing the Spectral branch reduces the ability to exploit inter-band correlations, which may cause weaker semantic discrimination in regions with similar spatial appearance. Without FEM, edge and texture details become less distinct, indicating that frequency-domain enhancement is useful for compensating degraded infrared details. Removing DGM results in more obvious structural discontinuities and less accurate local regions, especially around objects with geometric variations. This shows that DGM plays an important role in refining local structures and suppressing redundant background responses. The internal comparison of DGM further suggests that DCN contributes to deformation-aware local modeling, while ASM helps emphasize informative structures and reduce background redundancy.

These observations are further supported by \autoref{fig:Ablation_DGM_FEM}. DGM helps suppress background clutter and highlight structurally important regions, such as vehicles and buildings, whereas FEM enhances high-frequency cues associated with edges and textures. Their combination improves local structural consistency and semantic reliability, leading to more coherent colorization results in complex road scenes.

\section{Conclusion}

This paper investigates infrared hyperspectral image colorization from the perspective of coordinated spatial, spectral, and frequency-domain representation. The proposed FSCM framework combines state-space modeling with frequency-aware detail enhancement, enabling efficient long-range dependency modeling while compensating for the weak textures and blurred boundaries commonly observed in infrared imagery. By further introducing local geometry-adaptive refinement and semantic-level supervision, the method improves structural preservation, color naturalness, and semantic consistency in challenging road and remote-sensing scenes. Experimental results on multiple datasets show that FSCM achieves better visual quality and quantitative performance than representative infrared colorization methods, while maintaining manageable computational complexity. These results suggest that frequency-enhanced spatial–spectral state-space modeling provides an effective solution for infrared hyperspectral image colorization and offers a promising direction for future infrared visual perception tasks.

\bibliographystyle{elsarticle-num}
\bibliography{cas-refs}

\end{document}